\documentclass[sigconf, nonacm]{acmart}

\usepackage{url}
\usepackage[ruled,vlined]{algorithm2e}
\usepackage{pgfplots}
\usepackage{multirow}
\usepackage{dcolumn}
\usepackage{graphicx}
\usepackage{subcaption}
\usepackage{hyperref}
\usepackage{float}
\usepackage{algpseudocode}
\usepackage{amsmath}
\usepackage{balance}
\usepackage[a-1b]{pdfx}
\usepackage{hyperref}
\usepackage{graphicx}
\usepackage{listings}
\usepackage{pythonhighlight}
\usepackage{listings}
\usepackage{hyperref}
\usepackage{url}

\hypersetup{
    colorlinks=true,
    linkcolor=blue,
    filecolor=magenta,      
    urlcolor=cyan,
    pdftitle={Overleaf Example},
    pdfpagemode=FullScreen,
    }
\usepackage{xcolor}

\definecolor{codegreen}{rgb}{0,0.6,0}
\definecolor{codegray}{rgb}{0.5,0.5,0.5}
\definecolor{codepurple}{rgb}{0.58,0,0.82}
\definecolor{backcolour}{rgb}{1.0,1.0,1.0}

\begin{document}
\title{\textsc{SWIFT}:A Scalable lightWeight Infrastructure for Fine-Tuning}

\author{Yuze Zhao \quad Jintao Huang \quad Jinghan Hu \quad Xingjun Wang \quad Yunlin Mao \quad Daoze Zhang \quad Hong Zhang \quad Zeyinzi Jiang \quad Zhikai Wu \quad Baole Ai \quad Ang Wang \quad Wenmeng Zhou \quad Yingda Chen}
\affiliation{%
  \institution{ModelScope Team, Alibaba Group \\ 
  \small{\texttt{\{yuze.zyz,huangjintao.hjt,xingjun.wxj,maoyunlin.myl,zhangdaoze.zdz,hujinghan.hjh,zh461848,zeyinzi.jzyz,wuzhikai.wzk,\\ 
  baole.abl,wangang.wa,wenmeng.zwm,yingda.chen\}@alibaba-inc.com}} \\
  }
  \city{HangZhou}
  \country{China} \\
  \href{https://github.com/modelscope/ms-swift}{\textbf{https://github.com/modelscope/ms-swift}}
}

\begin{abstract}
Recent development in Large Language Models (LLMs) and Multi-modal Large Language Models (MLLMs) have leveraged Attention-based Transformer architectures and achieved superior performance and generalization capabilities. They have since covered extensive areas of traditional learning tasks. For instance, text-based tasks such as text classification and sequence labeling, as well as multi-modal tasks like Visual Question Answering (VQA) and Optical Character Recognition (OCR), which were previously addressed using different models, can now be tackled based on one foundation model. Consequently, the training and lightweight fine-tuning of LLMs and MLLMs, especially those based on the Transformer architecture, have become particularly important. In recognition of these overwhelming needs, we develop SWIFT, a customizable one-stop infrastructure for large models. With support of more than $550+$ LLMs, $200+$ MLLMs, and almost 200 Megatron models, SWIFT stands as the open-source framework that provides the \textit{ most comprehensive support} for fine-tuning large models. In particular, it is the first training framework that provides systematic support for MLLMs.  In addition to the core functionalities of fine-tuning, SWIFT also integrates post-training processes such as inference, evaluation, and model quantization, to facilitate fast adoption of large models in various application scenarios.  With a systematic integration of various training techniques, SWIFT offers helpful utilities, such as benchmark comparisons among different training techniques for large models. For fine-tuning models specialized in agent framework, we show that notable improvements on the ToolBench leaderboard can be achieved by training with customized datasets on SWIFT, with an increase of 5.2\%-21.8\% in the Act.EM metric over various baseline models, a reduction in hallucination by 1.6\%-14.1\%, and an average performance improvement of 8\%-17\%.
\end{abstract}

\maketitle

\section{Introduction}
In the past few years, the Transformer \cite{NIPS2017_3f5ee243} has been widely recognized as one of the dominant architectures for large models. In the early stages,  encoder-only structures are utilized to accomplish tasks such as text classification and sequence labeling, with models like BERT\cite{devlin2018bert} serving as typical examples. Conversely, encoder-decoder and decoder-only structures were used mostly for text generation tasks. In comparison, vision models often adopts ResNet architecture to handle tasks like Visual Question Answering (VQA), Object Detection, and Image Segmentation. These early approaches to deep learning tasks were characterized by the use of distinct model structures for different tasks.

The abundance in computational power and structured training data has brought out potentials of Transformer-based models, as industry begins to overtake models tuned for single-task. This shift positions Transformer as the preferred architectures for open-domain applications. Most notable examples include the GPT models\cite{NEURIPS2020_1457c0d6,radford2019language} from OpenAI, as well as M6 \cite{lin2021m6} and OFA \cite{pmlr-v162-wang22al} models. Such progress highlighted the feasibility of using a single model to address multiple closed-domain tasks. Consequently, the paradigm of leveraging large-scale foundation models for generative tasks has emerged as the new standard for addressing multiple tasks including text classification and sequence labeling. The attention mechanism has also gained traction in addressing distinct multi-modal tasks with one foundation model.  The release of Qwen2.5-VL \cite{bai2025qwen25vltechnicalreport}, GLM4-V \cite{glm2024chatglm}, InternVL3 \cite{zhu2025internvl3exploringadvancedtraining}, and DiT models \cite{peebles2023scalablediffusionmodelstransformers} , can all satisfy to this trend. These foundation models exhibit robust capabilities in open-domain image-text and video-text question answering, as well as in image generation. They have also shown potentials for recognizing detailed image information and performing bounding-box annotation, achieving results comparable to previous closed-domain solutions.

Throughout the development of large models, open-source communities have played a critical role. Platforms such as Hugging Face\footnote{Web: https://huggingface.co/; GitHub: https://github.com/huggingface}, and ModelScope\footnote{Web: https://modelscope.cn; GitHub: https://github.com/modelscope} are both notable examples in promoting sharing and development of large models. Launched in 2017, Hugging Face was initially tasked with the mission to address issues related to the PyTorch version of BERT. The resulting Transformers library later became the \textit{de facto} standard for implementing large models. At the same time, the Hugging Face trainer also supports multi-node and multi-GPU parallel training methods such as DeepSpeed and FSDP, making it one of the most widely used trainers. 
For alignment techniques, TRL\footnote{GitHub:https://github.com/huggingface/trl} is introduced to extend base trainer class from Transformers and implements specific methods to support techniques such as Direct Preference Optimization (DPO)\cite{rafailov2024directpreferenceoptimizationlanguage}, Optimized Reward Policy Optimization (ORPO)\cite{hong2024orpomonolithicpreferenceoptimization}, and Knowledge Transfer Optimization (KTO)\cite{ethayarajh2024ktomodelalignmentprospect}.

Reinforced fine-tuning(RFT) has consistently served as a crucial mechanism for enhancing model performance following Supervised Fine-Tuning. The primary methodology involves utilizing data generated either by the model itself or by alternative models, which is subsequently filtered through reward models or reward functions, and then employed for continued model training. Among the commonly utilized reinforced fine-tuning approaches are Rejection sampling Fine-Tuning\cite{yuan2023scalingrelationshiplearningmathematical} and distillation. Following DeepSeek's introduction of the DeepSeek-R1\cite{deepseekai2025deepseekr1incentivizingreasoningcapability} model, RLVR methods, especially represented by Generalized Reinforcement Policy Optimization (GRPO)\cite{shao2024deepseekmathpushinglimitsmathematical} emerged as a significant reinforced fine-tuning technique. The principal advantage of this approach lies in its reinforcement learning characteristics, which enable substantial improvements in metrics such as Pass@K without necessitating massive SFT data (requiring only minimal data for code initialization). In response to these technological advancements, the open-source community has developed numerous RL frameworks, including Open-R1\footnote{GitHub: https://github.com/huggingface/open-r1}, veRL\footnote{GitHub: https://github.com/volcengine/verl}, and OpenRLHF\footnote{GitHub: https://github.com/OpenRLHF/OpenRLHF}.

Given the large number of parameters and high memory consumption of large models, their out-of-box training has become as a significant bottleneck in the proliferation of AI. Early solutions, such as Prefix Tuning \cite{li2021prefixtuning}, Prompt Tuning \cite{lester-etal-2021-power}, and P-Tuning \cite{liu2021gpt, DBLP:journals/corr/abs-2110-07602} open the chapter of resource-efficient training, but they can suffer from  ``knowledge forgetting'' -- a phenomenon description the scenario where fine-tuned LLMs may lost it general capacities from the foundation models. The introduction of LoRA \cite{hu2022lora} shows the potentials of reduces memory consumption during training significantly comparing to full-parameter training, without loss of generality in models. This allows developers to embark on efficient-training on  domain data using hardware that is much easier to access. Subsequently, more similar techniques were introduced, such as the enhancement algorithm rsLoRA \cite{kalajdzievski2023rankstabilizationscalingfactor}, DoRA \cite{liu2024doraweightdecomposedlowrankadaptation}, PISSA \cite{meng2024pissaprincipalsingularvalues}, OLoRA \cite{buyukakyuz2024oloraorthonormallowrankadaptation}, LoRA+ \cite{hayou2024loraefficientlowrank}, and the LLaMA-Pro \cite{wu2024llamaproprogressivellama} have proliferated, providing an array of new techniques for efficient fine-tuning. In recognition of the vast differences among these different techniques, efforts begin to emerge to unify training interfaces. For example, Hugging Face has introduced the PEFT \footnote{GitHub: https://github.com/huggingface/peft} that specializes in collecting and standardizing interfaces for efficient fine-tuning algorithms. 


In addition to lightweight training techniques based on additional structures such as LoRA, quantization stands as another solution for reducing memory consumption during training. Typically, LLMs use 16-bit half-precision formats, such as float16 and bfloat16, for both inference and training. By reducing the tensor types to 8-bit or 4-bit, the same model can be loaded with less memory. It is even possible to run the model with 1-bit or 1.5-bit precision; this approach is collectively known as quantization. For instance, BitsAndBytes \cite{dettmers2022llmint8} employs segmented quantization with outlier thresholds, AutoGPTQ \cite{frantar2023gptqaccurateposttrainingquantization} performs Taylor series decomposition on parameter matrices and uses the Hessian matrix to evaluate parameter importance, and AWQ \cite{lin2024awqactivationawareweightquantization} evaluates parameter significance and applies scaling factors for quantization. Due to the complexity of quantization techniques and their poor adaptability to different models, the Hugging Face community has introduced the Optimum library\footnote{GitHub: https://github.com/huggingface/optimum} as a unified implementation for various quantization methods. Nevertheless, the task of LLM training and fine-tuning remains formidable for most developers, as the aforementioned solutions only cover support of a limited number of models and techniques. In particular, support for newer models and techniques are often lacking  in existing solutions. Furthermore, to ensure that the trained models can be effectively deployed, the post-training processes, such as inference and evaluation, are also steps in utilization of trained models. To address this, we introduced SWIFT \footnote{GitHub:https://github.com/modelscope/swift}, an open-source framework targeted at facilitating lightweight training of large models, which also incorporates comprehensive functionality for post-training processes. SWIFT assists developers to perform training and inference operations with minimal learning overhead. By streamlining various technical components, sourced or self-developed, in a unified way, SWIFT is tasked to enable efficient training and development pipelines of large models.

Specifically, our contributions can be summarized as follows:

\begin{itemize}
\item We introduce SWIFT, a training framework compatible with the general model standards of the Transformers library. SWIFT integrates libraries such as PEFT and Optimum, enabling pre-training, fine-tuning, human alignment and reinforced fine-tuning for LLM and MLLM models. In addition quantization training (QLoRA \cite{dettmers2023qloraefficientfinetuningquantized}) is supported as well. Today SWIFT supports over 550 LLM models and over 200 MLLM models, encompassing all major open-source models. It also comes with support of over 150 pure text and multi-modal datasets. 

\item Other than the standard Attention structures, training and inference for model structures such as Mamba \cite{gu2024mambalineartimesequencemodeling} model are also supported in SWIFT. Training with Megatron \cite{shoeybi2020megatronlmtrainingmultibillionparameter} structured models is supported as well which facilitates large-scale parallel pre-training across multiple nodes and GPUs to be performed with SWIFT.

\item Several SOTA tuners are implemented or planted in SWIFT project to enhance lightweight training. These tuners are designed to be used independent of our SWIFT training loop to allow more flexible usage.

\item Numerous post-training operations are integrated in SWIFT library, including quantization (BNB/GPTQ/AWQ, etc.), LoRA merging, evaluation (supporting over 100+ pure text and multi-modal evaluation sets), as well as inference and deployment capabilities. For deployment, we support  native PyTorch deployment and inference acceleration based on vLLM \cite{kwon2023efficient} , and LMDeploy \cite{2023lmdeploy}, together these integrations provide support for inference against most text and multi-modal large models.
\end{itemize}

In summary, we have comprehensively constructed a complete technical chain around LLM training, effectively reducing the cost of understanding and using large models. Particularly for the training of multi-modal models, to our knowledge, we are the first open-source framework to establish a comprehensive multi-task training and complete end-to-end solution for numerous multi-modal large models.

\begin{table*}[ht]
    \centering
    \begin{tabular}{lcccccc}
    \hline
    & LLaMA-Factory & FireFly   & FastChat &  Axolotl&  LMFlow&  SWIFT(Ours)\\
    \hline
     LoRA & \checkmark & \checkmark  & \checkmark & \checkmark & \checkmark & \checkmark \\
     QLoRA  & \checkmark & \checkmark  & \checkmark & \checkmark & \checkmark & \checkmark \\
      LLaMA-Pro & \checkmark &   &   &   &   & \checkmark \\
       LongLoRA\cite{chen2024longloraefficientfinetuninglongcontext} & \checkmark & \checkmark &   & \checkmark &    & \checkmark \\
       GaLore\cite{zhao2024galore} & \checkmark &   &   & \checkmark &   & \checkmark \\
   Q-GaLore\cite{zhang2024qgalore} &  &    &   & \checkmark &   & \checkmark \\ 
   FourierFt\cite{gao2024parameterefficientfinetuningdiscretefourier} & &    &   &   &   & \checkmark \\
   LoRA+ & \checkmark &    &   & \checkmark &   & \checkmark \\
   LISA & &    &   & \checkmark & \checkmark & \checkmark \\
   DoRA & \checkmark &    &   & \checkmark &   & \checkmark \\
   rsLoRA & \checkmark &    &   & \checkmark &   & \checkmark \\
   UnSloth & \checkmark & \checkmark  &   & \checkmark &   & \checkmark \\
   \hline
    LLM-PRETRAIN & \checkmark & \checkmark & \checkmark & \checkmark & \checkmark & \checkmark \\
    LLM-Megatron-PRETRAIN & &   &   &   &   & \checkmark \\
    LLM-SFT & \checkmark & \checkmark & \checkmark & \checkmark & \checkmark & \checkmark \\
    LLM-DPO & \checkmark & \checkmark  &   & \checkmark & \checkmark & \checkmark \\
    LLM-CPO\cite{xu2024contrastivepreferenceoptimizationpushing} & &   &   & \checkmark &  & \checkmark \\
    LLM-ORPO & \checkmark &   &   & \checkmark &    & \checkmark \\
    LLM-KTO  & \checkmark &   &   & \checkmark &    & \checkmark \\
    LLM-SimPO\cite{meng2024simposimplepreferenceoptimization} & \checkmark &   &   & \checkmark &    & \checkmark \\
    \hline
    MLLM-PRETRAIN & 60+ models &    &   &  20+ models &   & 200+ models \\
    MLLM-SFT & 60+ models &    &   & 20+ models  &  & 200+ models \\
    MLLM-RLHF & 60+ models &    &   & 20+ models  &  & 200+ models \\
    \hline
    vLLM & \checkmark &   & \checkmark &   & \checkmark  & \checkmark \\
    LMDeploy & &   &   &   &    & \checkmark \\
    LLM Evaluation & 3 datasets &   & \checkmark &   & \checkmark  & 48 datasets, 2 custom datasets \\
    MLLM Evaluation  & &   &   &   &    & 95 datasets \\
    WEB-UI & \checkmark &   & \checkmark &   &    & \checkmark \\ 
    \hline
    \end{tabular}
    \caption{The comparison of support for training auxiliary capabilities}
    \label{tab:c1}
\end{table*}

\section{Related Works}
LLaMA-Factory\cite{zheng2024llamafactoryunifiedefficientfinetuning} is a versatile, all-in-one large model training framework. This framework is fully compatible with the Hugging Face model ecosystem. Additionally, it supports a WEB-UI based on Gradio, further reducing the cost of usage. LLaMA-Factory supports the pre-training, fine-tuning, and human alignment of over 100 text LLMs. It also facilitates the training of some of them multi-modal models such as LLaVA, PaliGemma, and Qwen-VL. In terms of evaluation capabilities, it supports the evaluation processes for the CEVAL, MMLU, and CMMLU datasets and enables inference and deployment workflows based on vLLM.

Firefly\footnote{GitHub:https://github.com/yangjianxin1/Firefly} leverages the transformers training ecosystem (Trainer/PEFT, etc.). Remarkably, it explores training datasets and creates several popular datasets for NLP training, such as firefly-train-1.1M, moss-003-sft-data, and ultrachat. They have also utilized these dataset for lightweight training on various models, including firefly-mixtral-8x7b, which has outperformed Yi-34B-Chat on multiple leaderboards.

FastChat\cite{zheng2023judging} is a comprehensive training and inference framework. This framework is equipped with capabilities for model training, inference, deployment, and evaluation. FastChat has leveraged Transformers and PEFT for training, and supports models such as LLaMA, Vicuna, and T5. It supports lightweight fine-tuning using techniques such as LoRA, QLoRA, and XFormers\footnote{GitHub: https://github.com/facebookresearch/xformers}. For deployment, FastChat supports inference acceleration frameworks like vLLM, SGLang\footnote{https://github.com/sgl-project/sglang}, and LightLLM\footnote{https://github.com/ModelTC/lightllm}. FastChat places a focus on model inference and deployment, with relatively limited training support.

Axolotl\footnote{https://github.com/axolotl-ai-cloud/axolotl} employs training component libraries such as TRL, PEFT, and Transformers. This framework has extended training capabilities, including the encapsulation of the mambassm\footnote{GitHub: https://github.com/state-spaces/mamba} library, thereby enhancing the ability to train these models using the transformers ecosystem. Axolotl supports the LoRA and QLoRA training of various models across multiple series, including LLaMA, Mistral, Qwen, and Phi, and also supports inference and merge-LoRA operations.

The LMFlow\cite{diao2023lmflow}  encapsulates  model training process in a pipeline style. It supports SFT and RLHF training for LLM models like LLaMA, Gemma, and Qwen, and allows for custom optimizers and tuners, such as LoRA. Additionally, it has developed lightweight fine-tuning techniques like LISA\cite{pan2024lisalayerwiseimportancesampling}. LMFlow provides capabilities for evaluating LLMs and supports inference and inference acceleration for both pure text and multi-modal models.

We have summarized  capabilities of all these frameworks in the table \ref{tab:c1} for easy reference.

\section{Implementations and Frameworks}

We believe that unifying multiple model architectures to enhance the all-around capability of a model, is an important trend in the development of large models. For instance, the primary distinction between text and multi-modal LLMs lies in the additional vision-tower component. The hidden states from this vision-tower, once processed through a projector, are integrated into the LLM's embeddings. Furthermore, the majority of multi-modal models can support text input and perform inference in the manner of a text-only model. 

Pre-training a text-only model typically requires processing data that amounts to the order of tens of terabytes of tokens, which is fastly approaching the limit of exhausting all effective text corpus available. However, from a multi-modal perspective, the hidden states of text, image, and video data can be interchanged in high-dimensional space\cite{radford2021learningtransferablevisualmodels}. Consequently, models trained through multi-modal pre-training could be considered to possess virtually unlimited data, therefore they will exhibit a significant data advantage over those trained solely on text. To this end, it is our belief that multi-modal models shall become predominant in future model development. In our framework design, we strive to eliminate the gap between training pure text LLMs and multi-modal LLMs, and we do this by establishing unified standards in data processing, model templates, and model training.

Training, or fine-tuning is not the end of LLM applications. Once a model is trained, there is often needs for convenient and efficient evaluation processes to determine model quality. These evaluation processes can even be integrated into the training phase for cross-validation (e.g., incorporating gsm8k evaluation during training), or during inference to conduct comprehensive evaluations on specific datasets. Additionally, post-training quantization of models can be performed to use quantized models for service, ensuring minimal memory usage while maintaining theoretical performance. 

This necessity applies to model deployment as well. Efficient inference and deployment of various post-training checkpoints, including original models, LoRA models, LLaMA-Pro models, and quantized models, are equally important. Therefore, integrating upstream and downstream capabilities within the framework, alongside the training itself is crucial. This integration will not only streamline the overall process of model application, but also enable exploration different capabilities together. The joint relationship can be found between evaluation and deployment,  between evaluation and training, as well as between quantization and deployment. We believe to truly lower the barrier for model utilization, it is of paramount importance to construct a unified framework for both text and multi-modal LLMs that centered around training capabilities and its downstream integration.

\begin{figure}[ht]
    \centering
    \includegraphics[width=0.5\textwidth]{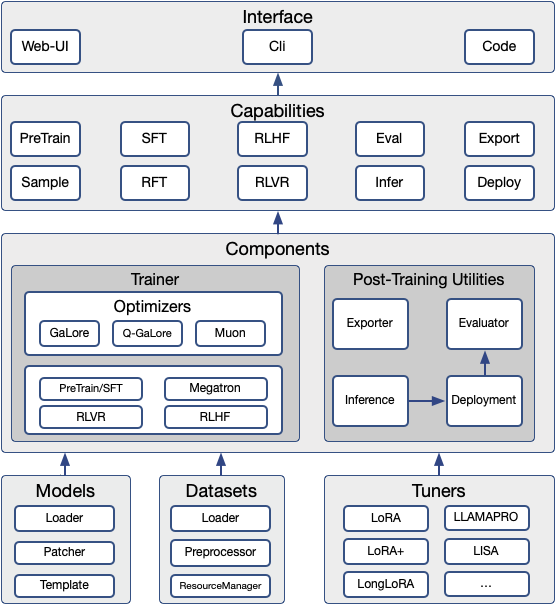}
    \caption{The framework of SWIFT}
    \label{fig:framework}
\end{figure}

\subsection{Design of Training Architecture}

SWIFT supports several categories of lightweight tuning techniques:

\begin{itemize}
     \item Reducing Trainable Parameters: This involves partially training the original model's parameters. Reducing the number of trainable parameters can effectively decrease the number of gradient values. For instance, LISA, which randomly activates different layers, can significantly reduce memory usage without notably decreasing training accuracy.
    \item Model Quantization: Quantization is a crucial method for reducing memory pressure. The main idea is to convert the low-precision floating-point values of the model into 8-bit or lower fixed-point values. SWIFT currently supports six types of quantization: BNB, HQQ, EETQ, AWQ, GPTQ, and AQLM. Generally, quantization methods are combined with additional structures for training, such as QLoRA.
     \item Reducing Memory Usage of Gradient Values: Techniques such as GaLore perform SVD decomposition on gradient values, effectively reducing the memory required for storing these values.
     \item Freezing the Original Model: This approach supports training with additional structures. Typical implementations include LoRA and AdaLoRA. 
     \item Sharding or Mixed Precision: Examples include DeepSpeed Zero1/2/3, FSDP, and mixed precision training.
\end{itemize}

As show in Fig. \ref{fig:framework}, tuners can leverage and extend the capabilities of the PEFT library. For instance, tuners incorporate techniques such as LoRA, AdaLoRA \cite{zhang2023adaloraadaptivebudgetallocation}, IA3 \cite{liu2022fewshotparameterefficientfinetuningbetter}, BOFT \cite{liu2024parameterefficientorthogonalfinetuningbutterfly}, and Vera \cite{kopiczko2024veravectorbasedrandommatrix}. These tuners are introduced with adjustments to ensure compatibility and seamless operation within MLLMs during training. Additionally, SWIFT offers support for a much wider range of tuners when comparing with PEFT, including SCEdit \cite{jiang2023sceditefficientcontrollableimage} and ResTuning \cite{jiang2023restuningflexibleefficienttuning}, as well as LLaMA-Pro, LongLoRA, and LISA. These tuners can be used in combination, similar to the MixedPeftModel capability of PEFT, and support offloading of deactivated tuners to CPU or meta devices. This  integration of tuners allows them to be applied to not only models supported within SWIFT, but also external models as well. SWIFT provides seamless support for both PEFT tuners and  customized tuners through its prepare\_model and from\_pretrained methods.

In the model functionality module, SWIFT provides a basic model loader that allows for flexible customization of model configurations. Given that various compatibility issues may arise during training, such as dtype conversion errors or tensor in-place change errors, SWIFT utilizes a patcher module to address these issues post model-loading, ensuring smooth operation in different scenarios including single-GPU, multi-GPU, full-parameter, or LoRA training.

In the dataset module, three types of data sources are supported. The first is MsDataset that loads dataset from ModelScope. The second is the `datasets` module from Hugging Face, which provides loading capabilities for Hugging Face datasets. Lastly, we support user-defined datasets, such as local CSV or JSONL files. A key feature of the dataset module is the pre-processing capability, which serves two main functions: converting different datasets into a standard format. The specific format details can be found in the appendix section\ref{sec:appendix}.

One of the critical components of the model module is the template. This component ensures that various models supported by SWIFT can correctly produce key fields such as input\_ids, attention\_masks, pixel\_values, and labels according to the design of model training. This module interfaces with the aforementioned standard dataset formats and converts these formats into different inputs as per the requirements of different models. Specifically, for multi-modal grounding tasks, bounding box (bbox) coordinates are converted within the template. For example, a bbox\_type 'real' represents actual pixel values, 'norm\_1000' represents values in thousandths, and 'norm\_1' represents normalized coordinate values. The template converts the actual coordinate values of the data into the coordinate values required by the model.

In the training component of the model, a significant part is the trainer, which includes both the SFT/PT trainer and the human alignment trainer. The former directly inherits from the trainer of Transformers and is used for predicting and training the cross-entropy of the next token. The latter inherits from the corresponding class of TRL and is used for training various RLHF algorithms such as DPO, ORPO, and KTO. For RLHF tasks of multi-modal models, we have made additional modifications and adaptations to ensure that all multi-modal models we support can use any compliant alignment dataset for RLHF training.

Specifically, in the direction of pre-training, SWIFT supports Megatron architecture models. Particularly in the CPT scenario, SWIFT first converts the checkpoints of the transformers architecture into the checkpoints of the Megatron architecture and then continues pre-training the model using various parallel methods of Megatron. After training, the checkpoints can be converted back to the format supported by transformers. 

In the training component, new optimizers such as GaLore, Q-GaLore and Muou are integrated, making them readily available for use during training. To further alleviate training pressure, SWIFT supports sequence parallelism technology \cite{jacobs2023deepspeedulyssesoptimizationsenabling}, which distributes sequences across different processes under DDP conditions, thereby reducing the memory consumption for long-sequence training.

Reinforced fine-tuning has gained prominence through notable works such as Rejection sampling Fine-Tuning, ReST\cite{gulcehre2023reinforcedselftrainingrestlanguage} and B-STaR\cite{zeng2025bstarmonitoringbalancingexploration}. This type of training methodology employs the model itself for rollout generation, followed by filtering correct responses via Outcome Reward Modeling (ORM), and subsequently retraining the model with the filtered data. This process can be iterated multiple times until model performance reaches a plateau. Empirical evidence demonstrates that Reinforced fine-tuning can improve model metrics by more than 5 points average following Supervised Fine-Tuning (SFT), without inducing significant knowledge forgetting across other evaluation benchmarks.

Similarly, the process of training a smaller model using rollout data from larger Language Models (LLMs) is generally referred to as distillation. This approach also incorporates ORM for data quality assessment. Following the emergence of Process Reward Modeling (PRM), both Rejection sampling and distillation methodologies have been enhanced by incorporating process rewards for superior data filtration. Regarding sampling techniques, various approaches including standard sampling, Diverse verifier tree search (DVTS), or Monte Carlo Tree Search (MCTS) can be employed. To facilitate this process, SWIFT provides the `swift sample` command, which supports methods such as vanilla sampling, MCTS, and DVTS. This command enables data sampling using open-source models and also supports sampling via closed-source APIs (distillation). Following the sample command, training can be conducted using either the `swift sft` or `swift rlhf` commands.

Generative Reinforcement Policy Optimization (GRPO) has emerged as a crucial method for enhancing model capabilities due to its minimal data requirements and high robustness. This has led to the establishment of a training pipeline consisting of pre-training, cold-start SFT, and RLVR(GRPO). GRPO is predominantly utilized for eliciting Chain-of-Thought (CoT) reasoning capabilities. For instance, format rewards can be implemented to compel models to engage in explicit reasoning (e.g., requiring the response to include paired `<think></think>` tags), followed by the provision of the correct answer (within `<result></result>` tags). GRPO has demonstrated substantial impact in improving models' Pass@K performance. Currently, GRPO is widely applied to enhance LLMs' CoT capabilities, as well as multi-modal LLMs' Visual Question Answering (VQA) logical reasoning and complex agent pathway capabilities. In response to this trend, SWIFT supports GRPO training methodologies(along with DAPO\cite{yu2025dapoopensourcellmreinforcement}) for both LLMs and MLLMs, and accommodates multi-round rollouts (for complex agent training pathways). SWIFT's GRPO implementation can readily scale to training systems with over a hundred GPUs and support models of increasing parameter sizes.

\begin{figure}[ht]
    \centering
    \includegraphics[width=0.5\textwidth]{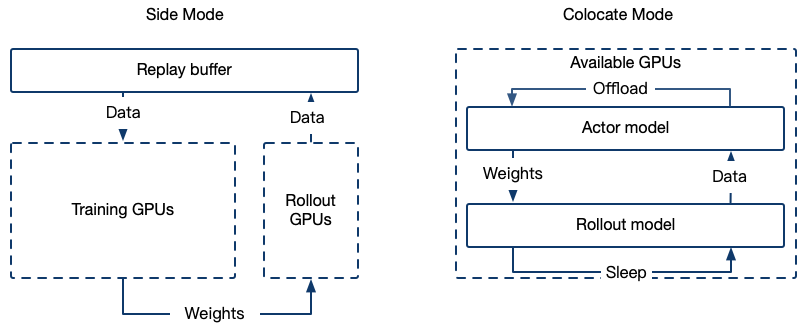}
    \caption{The GRPO model placement supported by SWIFT}
    \label{fig:grpo}
\end{figure}

To facilitate the use of LLM training, inference, and evaluation in actual production environments, we have released SWIFT on PYPI and supported various functionalities via the command line. For the training process, SWIFT can be easily invoked using command-line commands, which can be found in the appendix section\ref{sec:appendix}.

SWIFT provides several commands for different tasks: `pt` for pre-training, `sft` for fine-tuning, `rlhf` for RLHF, `sample` for sampling datasets. These commands are consistent for both pure text models and multi-modal models. For dataset selection, SWIFT supports the use of the `--dataset` option to directly use pure text and multi-modal datasets, and it also supports referencing local training files.

The highest-level interface is the web UI. For users who are familiar with graphical interfaces, using the web UI aligns more with their habits. The web UI uses Gradio as the foundational framework. Once the web UI is launched, users can select different training stages and adjust various training hyper-parameters directly in the interface. After the training starts, the interface will display training logs and loss/accuracy charts. For RLHF tasks, the charts will be replaced with metrics such as margin and logps that are relevant to the task type. This workflow is applicable to both pure text models and multi-modal models. Essentially, the SWIFT WEB UI serves as a command-line assembler. The web UI assembles the command-line strings for single-node multi-GPU or multi-node multi-GPU execution, and it uses these commands for multi-process background execution.

For lightweight training, SWIFT supports QLoRA training methods. The quantization methods available include BNB, HQQ \cite{badri2023hqq}, EETQ \cite{gordon2023eptqenhancedposttrainingquantization}, AWQ, GPTQ, and AQLM \cite{egiazarian2024extremecompressionlargelanguage}.  In terms of model support, we facilitate the training processes for over 300 NLP models and more than 50 multi-modal models. Specifically, to effectively fine-tune agents, we collaborated to create the MS-Agent dataset \footnote{WebPage: https://www.modelscope.cn/datasets/iic/ms\_agent}. This dataset is a relatively rare, high-quality Chinese fine-tuning dataset for agents. Subsequently, we updated the MSAgent-Pro dataset\footnote{WebPage: https://www.modelscope.cn/datasets/iic/MSAgent-Pro}, which adopts the ToolBench format. This dataset is very important for supervised fine-tuning (SFT) to enhance agent capabilities, and it includes the Chain of Thought (CoT) process, which significantly improves the effectiveness of multi-turn agent calls. To facilitate agent fine-tuning, SWIFT supports the `tools` field in dataset formats and allows fine-tuning training using different prompt formats (e.g., ToolBench \cite{qin2023toolllm} format, ReACT \cite{yao2023reactsynergizingreasoningacting} format, or other formats defined by model templates). This can be seen in the SWIFT standard dataset definitions.

SWIFT supports the `loss-scale` technique \cite{li2023modelscopeagentbuildingcustomizableagent}, which increases the training weight for important tokens. This makes it easier for the model to remember important content during learning. We used this technique to increase the weights for crucial parts of agent training, such as Action and Action\_Input fields, resulting in significant performance improvements compared to not using `loss-scale`.

In the multi-modal field, SWIFT provides comprehensive support, and various open-domain tasks can be run in SWIFT, such as Vision Question Answering (VQA), Optical Character Recognition (OCR), Grounded Captioning, and Referring Grounding.

\subsubsection{Design of Inference and Deployment Architecture}

Inference and deployment inherently possess a natural interdependence. The core logic of inference can be applied within deployment, and conversely, deployment can be viewed as a service encapsulation of inference. SWIFT's inference and deployment can be categorized into three types based on the backend: PyTorch Native(PT), vLLM, and LMDeploy. These three inference frameworks share identical parameters, allowing for the easy expansion of other inference acceleration frameworks in the future. One significant reason for SWIFT's encapsulation of inference for vLLM and LMDeploy is that, in cases where the original framework does not adequately support the model's templates, SWIFT can use its own templates to mask the differences between frameworks.

SWIFT employs FastAPI to encapsulate inference as a service, fully complying with the OpenAI universal interface definition\footnote{WebPage: https://platform.openai.com/docs/api-reference}. For the deployment of Agent capabilities, SWIFT supports OpenAI standard fields such as tools and tool, and it also supports inference and deployment of Agent data formats like ToolBench and ReACT in terms of data format. We have directly incorporated the concatenation of Agent prompts into the Template, which means that we can easily support the specific Agent formats of different models.

SWIFT supports the inference and deployment of both official and trained models, and these functionalities are equally supported in the WEB-UI. This implies that users can utilize SWIFT both as a deployment framework and as a ChatBot. Notably, we have integrated support for multi-LoRA inference and deployment in both the vLLM and PT backends. Specifically, users can conveniently switch between different LoRA configurations by specifying the respective LoRA names within the OpenAI interface, without the need to merge the LoRA models.

\subsubsection{Evaluation Architecture Design}

Evaluation and inference deployment are interdependent. The evaluation of models, particularly those that are post-training, depends on whether the models can initiate inference or deployment. In many evaluation frameworks, such as OpenCompass\cite{2023opencompass}, the standard OpenAI interface is directly used as a dependency, which is one of the reasons for supporting inference and deployment in SWIFT. In practice, developers may use different inference backends, such as vLLM or LMDeploy. Therefore, during evaluation, developers can flexibly choose different backends and deployment forms (e.g., official models, post-training LoRA models, post-training LLaMA-Pro models, merged models, quantized models) for evaluation, making the process more aligned with their actual use cases or ablation study scenarios.

To facilitate the use of custom datasets, SWIFT supports custom evaluation datasets for two types of tasks:
\begin{itemize}
    \item  Objective Question Evaluation Similar to CEval: Developers can format their datasets as CEval-style CSV files and then conduct evaluations, yielding conclusive results.
    \item Subjective Question Evaluation for QA: This evaluation uses standard metrics like ROUGE and BLEU. After writing the data into a CSV file, developers can perform evaluations.
    \item 
\end{itemize}
For evaluation capabilities, we rely on the EvalScope\footnote{GitHub:https://github.com/modelscope/evalscope} framework from the ModelScope Community. This framework constructs evaluation capabilities by integrating OpenCompass (for text models) and VLMEvalKit\cite{duan2024vlmevalkit} (for multi-modal models). By incorporating EvalScope, SWIFT currently supports over 100+ total pure text evaluation sets and multi-modal evaluation sets, as well as the aforementioned two types of custom evaluation datasets and their evaluation processes.

\subsubsection{Design of Quantization and Export Architecture}

The export module is primarily used for merging tuners, converting checkpoint formats, and quantization. Currently, the following types of export operations are supported within this module:
\begin{itemize}
    \item Merging Tuners: This includes merging tuners such as LoRA, LongLoRA, and LLaMA-Pro.
    \item Converting Checkpoints: This involves the mutual conversion of checkpoints between the Transformers format and the Megatron format.
    \item Quantization: At this stage, we support three quantization methods: AWQ, GPTQ, and BNB.
\end{itemize}

\begin{table}[ht]
\centering
\begin{tabular}{cccc}
\hline
\textbf{Quantize Method} & \textbf{QAT}& \textbf{QLoRA}& \textbf{PTQ}\\
\hline
BNB & \checkmark & \checkmark &  \checkmark\\
HQQ & \checkmark & \checkmark &  \\
EETQ & \checkmark & \checkmark &  \\
AWQ &  & \checkmark & \checkmark \\
GPTQ &  & \checkmark & \checkmark \\
AQLM &  & \checkmark & \\
\hline
\end{tabular}
\caption{Support for Quantization Methods}
\label{table:quantization_methods}
\end{table}

4. Exporting to Ollama: This process includes the incorporation of the model’s template configuration, allowing users to conveniently run the model using the `ollama` command.

\section{Experiments}

In addition to the algorithmic framework, we have also explored the tuning of the models and the underlying technology. Our objective is for SWIFT to serve not only as a framework but also as a way to validate the technology itself. To this end, we have divided our exploration of LLM training into several directions.

\subsection{Lightweight Tuning Benchmark}

We utilized SWIFT to replicate and validate the impact of various lightweight tuning algorithms on models. Using qwen-7b-chat as the base model, we conducted training on a single A100-80G GPU, comparing memory usage and loss, among other metrics.

\begin{table}[ht]
    \centering
    \begin{tabular}{cc}
        \hline
        \textbf{Hyper-parameter} & \textbf{Value} \\
        \hline
        batch\_size & 1 \\
        gradient\_accumulation\_steps & 16 \\
        epoch & 1 \\
        max\_length & 2048 \\
        learning\_rate & 5e-5 \\
        gradient\_checkpointing & true \\
        flash\_attn & true \\
        weight\_decay & 0.01 \\
        warmup\_ratio & 0.03 \\
        lora\_rank & 8 \\
        galore\_rank & 128 \\
        llamapro\_new\_blocks & 4 \\
        lisa\_activated\_layers & 2 \\
        \hline
    \end{tabular}
    \caption{Tuner benchmark hyper-parameter settings}
    \label{tab:hyper-benchmark}
\end{table}

The experiment hyper-parameters displays in chart \ref{tab:hyper-benchmark}, and the experiment results are summarized in table \ref{tab:swift-tuners}.

\begin{table}[ht]
    \centering
    \small
    \begin{tabular}{ p{1.5cm} p{1.4cm} p{2.1cm} p{1.2cm} p{1.2cm}}
        \hline
        \textbf{Tuner} & \textbf{Train/Eval loss} & \textbf{Trainable (M)} & \textbf{Memory (GiB)} & \textbf{Speed (samples/s)} \\
        \hline
        AdaLoRA   & 0.57 / 1.07  & 26.84 (0.35\%)  & 32.55  & 0.92 \\
        DoRA      & \textbf{0.53} / 1.01  & 19.25 (0.25\%)  & 32.46  & 0.51 \\
        \textbf{GaLore}    & 0.55 / 1.00  & 7721.32 (100\%) & 47.02  & 1.10 \\
        \textbf{Q-GaLore}  & 0.54 / 1.00  & 7721.32 (100\%) & 41.53  & 1.45 \\
        \textbf{LLaMAPro}  & \textbf{0.53} / 1.00  & 809.58 (9.49\%)  & 38.11  & 1.53 \\
        \textbf{LoRA+}     & \textbf{0.53} / 0.98  & 17.89 (0.23\%)  & 32.35  & 0.95 \\
        LoRA      & \textbf{0.53} / 1.01  & 17.89 (0.23\%)  & 32.35  & 0.95 \\
        RsLoRA    & \textbf{0.53} / 0.99  & 17.89 (0.23\%)  & 32.35  & 0.94 \\
        \textbf{LISA}      & 0.62 / 1.06  & -      & \textbf{31.11} & \textbf{2.66} \\
        Full      & 0.54 / \textbf{0.95}  & 7721.32 (100\%) & 73.53  & 1.43 \\
        \hline
    \end{tabular}
    \caption{\textmd{{Profiles of various Tuners}}}
    \label{tab:swift-tuners}
\end{table}

Among the benchmarks, "Full" represents the control group experiment using full-parameter training. It can be observed that the lowest memory consumption and fastest speed are achieved by LISA. Within the additional structure tuners, the lowest evaluation loss is recorded by LoRA+. In gradient reduction methods, Q-GaLore exhibits the lowest memory consumption. None of these tuning methods are included in the PEFT library.

\subsection{Agent Training}

Agent training constitutes an important category within model SFT. The quality of Agent training determines whether the model can be applied within an Agent framework to solve practical business problems. Generally, Agent capabilities are categorized into three types:

1. Document retrieval

2. Code Interpreter

3. API Calling

In the narrow sense, Agent training typically refers to the API Calling. This capability is positively correlated with the model's Chain of Thought (CoT) capability; the stronger the CoT capability, the better the model's understanding of APIs and its ability to reflect upon errors.

In this study, we utilized a mixed dataset comprising the ToolBench dataset and the AgentFlan \cite{chen2024agentflandesigningdatamethods} dataset, and conducted a series of experiments. We employed the LLaMA3-8b-instruct model and the Qwen2-7b-instruct model for training, and compared the results before and after training. The hyper-parameter settings are shown in the table \ref{tab:hyper-agent}.

\begin{table}[ht]
    \centering
    \begin{tabular}{cc}
        \hline
        \textbf{Hyper-parameter} & \textbf{Value} \\
        \hline
        batch\_size & 1 \\
        gradient\_accumulation\_steps & 32 \\
        epoch & 1 \\
        max\_length & 4096 \\
        learning\_rate & 2e-5 \\
        gradient\_checkpointing & true \\
        flash\_attn & true \\
        lora\_target\_modules & All linears \\
        lora\_rank & 8 \\
        \hline
    \end{tabular}
    \caption{Agent experiment hyper-parameter settings}
    \label{tab:hyper-agent}
\end{table}

In certain experiments, we employed the loss-scale technique to enhance the weights of some important tokens.

The ablation study comparing the loss-scale technique based on the LLaMA3-8b-instruct model is shown in the table \ref{table:loss-in-domain} and \ref{table:loss-out-of-domain}.

\begin{table}[ht]
\centering
\small
\begin{tabular}{cccccc}
\hline
\textbf{Model} & \textbf{Plan.EM} & \textbf{Act.EM} & \textbf{Hallu Rate} & \textbf{Avg.F1} & \textbf{R-L} \\
\hline
Original& 74.22 & 36.17 & 15.68 & 20.0 & 12.14 \\
w/o loss-scale & 84.29 & 55.71 & 4.85 & 49.40 & 25.06 \\
w/ loss-scale & \textbf{85.1} & \textbf{58.15} & \textbf{1.57} & \textbf{52.10} & \textbf{26.02} \\
\hline
\end{tabular}
\caption{LoRA (in-domain) ablation tests for loss-scale}
\label{table:loss-in-domain}
\end{table}

\begin{table}[ht]
\centering
\small
\begin{tabular}{cccccc}
\hline
\textbf{Model} & \textbf{Plan.EM} & \textbf{Act.EM} & \textbf{Hallu Rate} & \textbf{Avg.F1} & \textbf{R-L} \\
\hline
Original& 69.47 & 34.21 & 14.72 & 20.25 & 14.07 \\
w/o loss-scale & 85.10 & 55.55 & 5.26 & 48.52 & 31.22 \\
w/ loss-scale & \textbf{85.79} & \textbf{59.43} & \textbf{2.56} & \textbf{52.19} & \textbf{31.43} \\
\hline
\end{tabular}
\caption{LoRA (out-of-domain) ablation tests for loss-scale}
\label{table:loss-out-of-domain}
\end{table}

The experimental results indicate that introducing the loss-scale significantly improved all evaluation metrics.

The experimental results for Qwen2-7b-instruct are shown in the table \ref{table:qwen-in-domain} and \ref{table:qwen-out-of-domain}.

\begin{table}[ht]
\centering
\small
\begin{tabular}{cccccc}
\hline
\textbf{Model} & \textbf{Plan.EM} & \textbf{Act.EM} & \textbf{Hallu Rate} & \textbf{Avg.F1} & \textbf{R-L} \\
\hline
Original& 74.11 & 54.74 & 4.16 & 46.53 & 8.51 \\
GPT4& 80.28 & 55.52 & 5.98 & 48.74 & \textbf{28.69} \\
LoRA(Ours)& 77.05 & 56.97 & \textbf{0.9}& 49.53 & 19.81 \\
Full(Ours)& \textbf{83.37}& \textbf{60.01}& 2.58& \textbf{54.41}& 26.34\\
\hline
\end{tabular}
\caption{Qwen2-7b-instruct ToolBench (in-domain) results}
\label{table:qwen-in-domain}
\end{table}

\begin{table}[ht]
\centering
\small
\begin{tabular}{cccccc}
\hline
\textbf{Model} & \textbf{Plan.EM} & \textbf{Act.EM} & \textbf{Hallu Rate} & \textbf{Avg.F1} & \textbf{R-L} \\
\hline
Original& 73.17 & 57.67 & 3.84 & 48.58 & 11.23 \\
GPT4& 77.80 & 55.26 & 5.12 & 47.45 & 30.61 \\
LoRA(Ours)& 78.05 & 58.91 & \textbf{1.53}& 51.28 & 26.04 \\
Full(Ours)& \textbf{82.57}& \textbf{60.14}& 1.79 & \textbf{55.25}& \textbf{31.34}\\
\hline
\end{tabular}
\caption{Qwen2-7b-instruct ToolBench (out-of-domain) results}
\label{table:qwen-out-of-domain}
\end{table}

It can be observed that, compared to the official Qwen2 model, the average metrics after training improved by 8.25\%, and the model hallucinations were reduced to single digits. Moreover, most metrics surpassed those of GPT-4.

The experimental results for LLaMA3-8b-instruct are shown in the table \ref{table:llama-in-domain} and \ref{table:llama-out-of--domain}.

\begin{table}[ht]
\centering
\small
\begin{tabular}{cccccc}
\hline
\textbf{Model} & \textbf{Plan.EM} & \textbf{Act.EM} & \textbf{Hallu Rate} & \textbf{Avg.F1} & \textbf{R-L} \\
\hline
Original& 74.22 & 36.17 & 15.68 & 20.0 & 12.14 \\
LoRA(Ours)& \textbf{84.58} & \textbf{44.73} & \textbf{15.11}& \textbf{38.90} & \textbf{22.22} \\
\hline
\end{tabular}
\caption{Llama3-8b-instruct ToolBench (in-domain) results}
\label{table:llama-in-domain}
\end{table}

\begin{table}[ht]
\centering
\small
\begin{tabular}{cccccc}
\hline
\textbf{Model} & \textbf{Plan.EM} & \textbf{Act.EM} & \textbf{Hallu Rate} & \textbf{Avg.F1} & \textbf{R-L} \\
\hline
Original& 69.47 & 34.21 & 14.72 & 20.25 & 14.07 \\
LoRA(Ours)& \textbf{84.3} & \textbf{49.56} & \textbf{13.19}& \textbf{43.09} & \textbf{24.85} \\
\hline
\end{tabular}
\caption{Llama3-8b-instruct ToolBench (out-of-domain) results}
\label{table:llama-out-of--domain}
\end{table}

Based on LoRA training, the average metrics of LLaMA3 improved by 17\%. This demonstrates that open-source models and datasets are meaningful for Agent training in practical vertical scenarios. We have summarized the hyper-parameters and other experiences from the training process to facilitate replication and application by other developers. The mentioned dataset \footnote{https://www.modelscope.cn/datasets/iic/MSAgent-Pro} and  models \footnote{WebPage: https://modelscope.cn/models/swift/qwen2-7b-agent-instruct} \footnote{WebPage: https://modelscope.cn/models/swift/llama3-8b-agent-instruct-v2} can all be found on ModelScope.

\section{Conclusion}
In this paper, we described SWIFT, a lightweight, one-stop large model training framework from ModelScope. We hope that this framework can eliminate the mismatches between different models, datasets, and SOTA technologies, providing developers with a standardized solution that can solve the entire problem in a closed-loop manner. SWIFT supports over 300 LLMs and 50 MLLMs, and provides an easy-to-use WEB-UI based on the command line. Developers can perform various command-line operations on the WEB-UI, greatly reducing the cost of use. However, due to limited development time and other factors, SWIFT still has more features planned, such as:

1. Better support for Megatron large-scale parallel training. Currently, SWIFT's support for Megatron models does not fully cover mainstream LLMs and MLLMs. We hope that SWIFT can provide greater pre-training convenience for foundational model developers.

2. More in-depth multi-modal research. While SWIFT already supports training for most mainstream multi-modal models, we still lack more in-depth work on multi-modal datasets and models, such as providing high-quality datasets to prevent knowledge forgetting or training new multi-modal models using ModelScope’s self-developed datasets. Additionally, we hope to conduct more in-depth research on multi-modal Agents, multi-modal CoT, and multi-modal alignment training.

3. Support for RAG systems. We hope that SWIFT's training technology can be more SOTA and robust, making it easier to connect to various AI systems, such as enhancement training for RAG system models, helping RAG systems improve recall rates and answer accuracy.

\bibliographystyle{plain}
\bibliography{swift.bib}

\newpage
\appendix

\section{Supported Models and Datasets}
\label{sec:appendix}

\begin{table}[ht]
    \centering
    \begin{tabular}{ccc}
       \hline
       Supported Models  & Modal & Structure \\
       \hline
       LLaMA Series  & NLP  & Decoder-only \\
       Mistral/Mixtral Series  & NLP & Decoder-only \\
       Gemma Series  & NLP & Decoder-only \\
       Phi Series  & NLP & Decoder-only \\
       Qwen1/1.5/2 Series  & NLP & Decoder-only \\
       YI Series  & NLP & Decoder-only \\
       ChatGLM1/2/3 Series & NLP & Decoder-only \\
       DeepSeek1/2 Series & NLP & Decoder-only \\
       InternLM1/2 Series  & NLP & Decoder-only \\
       Mamba  & NLP & SSM \\
       PaliGemma Series  & Visual & Decoder-only \\
       Qwen-VL Series  & Visual & Decoder-only \\
       GLM4v  & Visual & Decoder-only \\
       DeepSeek-VL Series  & Visual & Decoder-only \\
       LLaVA Series  & Visual & Decoder-only \\
       InternVL1/2 Series  & Visual & Decoder-only \\
       Phi3-vision  & Visual & Decoder-only \\
       Yi-VL Series  & Visual & Decoder-only \\
       MiniCPM Series  & Visual & Decoder-only \\
       lorence Series  & Visual & Encoder-Decoder \\
       Qwen-Audio Series  & Audio & Decoder-only \\
       \hline
    \end{tabular}
    \caption{Part of models SWIFT supported}
    \label{tab:my_label}
\end{table}

\begin{table}[ht]
    \centering
    \begin{tabular}{cccc}
       \hline
       Supported Datasets  & Modal & Task & Language\\
       \hline
       alpaca-en & NLP & QA & English \\
       synthetic-text-to-sql & NLP & Text to Sql & English \\
       firefly-train-1.1M  & NLP  & QA & Chinese \\
       deepctrl-sft  & NLP  & QA & Chinese \\
       ruozhiba  & NLP  & QA & Chinese \\
       ms-agent & NLP & Agent & Chinese \\
       ms-agent-pro & NLP & Agent & English \\
       chinese-c4  & NLP  & Pretrain & Chinese \\
       fineweb & NLP  & Pretrain & Chinese \\
       okvqa/a-okvqa & Vision & VQA & English \\
       chart-qa & Vision  & VQA & English \\
       ocr-vqa & Vision  & OCR & English \\
       llava-pretrain & Vision  & VQA & English \\
       llava-instruct-150k & Vision  & VQA & English \\
       mantis-instruct & Vision  & VQA & English \\
       grit & Vision  & VQA & English \\
       science-qa & Vision  & VQA & English \\
       refcoco/refcocog & Vision  & Grounding & English \\
       rlaif-v & Vision  & RLHF & English \\
       aishell1-zh-mini & Audio & Audio QA & English \\
       \hline
    \end{tabular}
    \caption{Part of datasets SWIFT supported}
    \label{tab:my_label}
\end{table}

\newpage
\section{Loss-scale settings}
\label{sec:appendix2}

\begin{table}[ht]
    \centering
    \begin{tabular}{cc}
        \hline 
        \textbf{Pattern} & \textbf{Value} \\
        \hline 
        The response of tool selection query & 3.0 \\
        The response of param recalling query & 3.0 \\
        The response of param name query & 3.0 \\
        The response of param value query & 3.0 \\
        The content of 'Name:' & 3.0 \\
        The content of 'Action:' & 3.0 \\
        The content of 'Action Input:' & 3.0 \\
        The content of 'Tool:' & 3.0 \\
        The content of 'Command' & 3.0 \\
        The content of 'Arguments:' & 3.0 \\
        'Observation:' & 2.0 \\
        \hline
    \end{tabular}
    \caption{The weight of content in Agent training for loss-scale tesing}
    \label{tab:my_label}
\end{table}

\section{SWIFT commands}
\label{sec:appendix3}

\begin{figure*}[ht]
\centering
\begin{lstlisting}[language=Python, caption=The training and inference code of tuners, label=code:tuners, frame=single, numbers=left, keywordstyle=\color{blue}]
# Prepare a tuner
model = Swift.prepare_model(model, {'lora': LoRAConfig(), 
                                    'llamapro': LLaMAProConfig()})
# Load checkpoint
model = Swift.from_pretrained(model, 'some-training-ckpt-dir')

# Simple code for training
model = Model.from_pretrained('Qwen/Qwen3-8B'...)
model = Swift.prepare_model(model, 
            {'first_tuner': LoraConfig(...), 
             'second_tuner': LLaMAProConfig(...))
train_data = MsDataset.load('<dataset-id>', split='train')
eval_data = MsDataset.load('<dataset-id>', split='eval')    

trainer = Seq2SeqTrainer(
    model=model,
    args=Seq2SeqTrainingArguments(learning_rate=1e-4...),
    train_dataset=train_data, eval_dataset=eval_data)

trainer.train()
\end{lstlisting}
\end{figure*}

\begin{figure*}[ht]
\centering
\begin{lstlisting}[language=Python, caption=The standard prompts of SWIFT, label=code:dataset, frame=single, numbers=left, keywordstyle=\color{blue}]
# QA
{"query": "Calculate 22+45", "response": "The answer is 67."}
# QA with history and tools
{"system": "You are a good math teacher.", 
"query": "Calculate 22+45", 
"response": "The answer is 67.",
"history": [["Can you calculate math?", 
"Yes, I can do math calculation."]],
"tools": [{"type": "function", "function": 
{"name": "get_current_weather", ...]}, ...}
# RLHF
{"query": "Calculate 22+45", "response": "The answer is 67.", 
"rejected_response": "I cannot calculate math."}
# VQA
{"query": "<image>What is in the image?", 
"response": "The image shows a little girl walking.",
"images": ["/coco2017/train/10045.jpg"]}
# Multi-Modal RLHF
{"query": "<image>What is in the image?", 
"response": "The image shows a little girl walking.",
"rejected_response": "I cannot see any image.",
"images": ["/coco2017/train/10045.jpg"]}
# Grounding
{"query": "<image>Where is <ref-object>?", 
"response": "The position is <bbox>",
"images": ["/coco2017/train/10045.jpg"],
"objects": "[{\"caption\": \"guy in red\", 
\"bbox\": [138, 136, 235, 359], 
\"bbox_type\": \"real\", \"image\": 0}}
\end{lstlisting}
\end{figure*}

\begin{figure*}[ht]
\centering
\begin{lstlisting}[language=Python, caption=The SWIFT command lines, label=code:train, frame=single, numbers=left, keywordstyle=\color{blue}]
# Multi-GPU sft command
CUDA_VISIBLE_DEVICES=0,1,2,3,4,5,6,7 \
NPROC_PER_NODE=8 \
swift sft \
    --model Qwen/Qwen3-8B \
    --dataset AI-ModelScope/blossom-math-v2 \
    --deepspeed zero3
# Single GPU RLHF command
swift rlhf 
    --model Qwen/Qwen3-8B \
    --rlhf_type dpo \
    --dataset AI-ModelScope/hh-rlhf
# GRPO command
swift rlhf 
    --rlhf_type grpo \
    --model Qwen/Qwen3-8B \
    --dataset 'open-r1/verifiable-coding-problems-python-10k'
# Inference a multi-modal model
swift infer 
    --model Qwen/Qwen2.5-VL-3B-Instruct \
    --infer_backend vllm
# Deploy a checkpoint by vllm
swift deploy 
    --ckpt_dir /mnt/my-custom/ckpt-1100 \
    --infer_backend vllm
# Evaluate an nlp model
swift eval 
    --model Qwen/Qwen3-8B \
    --eval_dataset ceval gsm8k
# Evaluate a multi-modal model
swift eval 
    --model Qwen/Qwen2.5-VL-3B-Instruct \
    --eval_dataset COCO_VAL
# Evaluate an OpenAI url
swift eval 
    --eval_url https://127.0.0.1/8000 \
    --eval_dataset mmlu
# Evaluate use a custom dataset
swift eval 
    --model Qwen/Qwen3-8B \
    --custom_eval_config /mnt/my-dataset.json
# Merge LoRA
swift export --ckpt_dir /mnt/my-custom/ckpt-1100 --merge_lora true
# Quantize
swift export --ckpt_dir /mnt/my-custom/ckpt-1100 --quant_method awq
# Sample
swift sample 
    --model Qwen/Qwen3-8B \
    --sampler_engine vllm \
    --num_return_sequences 5 \
    --dataset AI-ModelScope/alpaca-gpt4-data-zh#5
# Distill
OPENAI_API_KEY="your_api_key" swift sample 
    --sampler_type distill \
    --model deepseek-r1 \
    --dataset tastelikefeet/competition_math#5
\end{lstlisting}
\end{figure*}

\end{document}